\def\year{2021}\relax
\documentclass[letterpaper]{article} 
\usepackage{comment}
\usepackage{booktabs} 
\usepackage[flushleft]{threeparttable}
\usepackage{subcaption}
\usepackage{aaai21}  
\usepackage{times}  
\usepackage{helvet} 
\usepackage{courier}  
\usepackage[hyphens]{url}  
\usepackage{graphicx} 
\usepackage{natbib}  
\usepackage{caption} 
\usepackage{xcolor}
\usepackage{caption} 
\usepackage{amsmath}
\usepackage{hyperref}
\usepackage{algorithm}
\usepackage{algorithmic}
\usepackage{mathtools}
\usepackage{multirow}
\usepackage{graphics}
\usepackage{verbatim}

\title{Explainable Deep Behavioral Sequence Clustering for Transaction Fraud Detection}

\author{
    Wei Min\textsuperscript{\rm 1}, Weiming Liang\textsuperscript{\rm 1}, Hang Yin\textsuperscript{\rm 1}, Zhurong Wang\textsuperscript{\rm 1}, Mei Li\textsuperscript{\rm 1}, Alok Lal\textsuperscript{\rm 2}\\
}

\affiliations{
    \textsuperscript{\rm 1}eBay China \hspace{4mm}
    \textsuperscript{\rm 2}eBay USA\\

}

\begin{document}

\maketitle

\begin{abstract}

In e-commerce industry, user-behavior sequence data has been widely used in many business units such as search and merchandising to improve their products.  However, it is rarely used in financial services not only due to its 3V characteristics -- i.e. Volume, Velocity and Variety -- but also due to its unstructured nature. In this paper, we propose a \textbf{Fin}ancial Service scenario \textbf{Deep} learning based \textbf{Behavior} data representation method for \textbf{Clustering} (\textbf{FinDeepBehaviorCluster}) to detect fraudulent transactions. To utilize the behavior sequence data, we treat click stream data as event sequence, use time attention based Bi-LSTM to learn the sequence embedding in an unsupervised fashion, and combine them with intuitive features generated by risk experts to form a hybrid feature representation. We also propose a GPU powered HDBSCAN (pHDBSCAN) algorithm, which is an engineering optimization 
for the original HDBSCAN algorithm based on FAISS project, so that clustering can be carried out on hundreds of millions of transactions within a few minutes. The computation efficiency of the algorithm has increased \textbf{500} times compared with the original implementation, which makes flash fraud pattern detection feasible. Our experimental results show that the proposed \textbf{FinDeepBehaviorCluster} framework is able to catch missed fraudulent transactions with considerable business values. In addition, rule extraction method is applied to extract patterns from risky clusters using intuitive features, so that narrative descriptions can be attached to the risky clusters for case investigation, and unknown risk patterns can be mined for real-time fraud detection. In summary, \textbf{FinDeepBehaviorCluster} as a complementary risk management strategy to the existing real-time fraud detection engine, can further increase our fraud detection and proactive risk defense capabilities.  

\end{abstract}


\maketitle

\section{Introduction}
User behavior analysis provides new insights into consumers' interactions with a service or product, many business units of e-commerce platforms rely on user behaviors heavily and to a great extent. For instance, search and merchandise are heavily driven by stochastic behaviors of users. However, user behavioral data is unstructured and sparse, it is rarely used in traditional financial services. User behavior describes the unique digital signature of the user, and is harder to fabricate, therefore brings opportunities to boost the capability of risk management. Recently, with the booming of deep learning, there is a growing trend to leverage user behavioral data in risk management by learning the representation of click-stream sequence. For example, e-commerce giants such as JD, Alibaba use recurrent neural network to model the sequence of user clicks for fraud detection\cite{session_based_fraud_jd,time_attention}, and Zhang et.al. \cite{sequence_online_loan} use both convolution neural network and recurrent neural network to learn the embedding of the click stream in online credit loan application process for default prediction. 

However, the common practice for risk management is to use a predictive framework, which is largely relying on feedback that is often lagged. According to Gartner Research,  "By 2021, 50\% of enterprises will have added unsupervised machine learning to their fraud detection solution suites",  quoted from Begin Investing now in Enhanced Machine Learning Capabilities for Fraud Detection. Unsupervised methods, especially clustering techniques are better suited to discover new types of unseen fraud. 
\begin{enumerate}
\item Fraud is a rare event, outlier detection framework provides a different angle to catch bad users that were missed by existing classification models;
\item Fraud is dynamic, supervised predictive learning can only help us learn existing fraud patterns, but unsupervised clustering is more capable of discovering unknown patterns;
\item Risk predictive models are usually trained on labeled data, with a performance tag from approved transactions. However, declined transactions also contain risk indicators and can be utilized in an unsupervised setting.
\end{enumerate}
Therefore, clustering techniques are effective complementary solutions to the existing risk predictive models. However, it can be argued that the outcome (the membership of data points) of the clustering task itself does not necessarily explicate the intrinsic patterns of the underlying data. From an intelligent data analysis perspective, clustering explanation/description techniques are highly desirable as they can provide interesting insights for pattern mining, business rule extraction and domain knowledge discovery. 

By combining the advantages of utilizing behavior sequential data and clustering techniques, we propose a framework called \textbf{FinDeepBehaviorCluster}: firstly, we use time-attention based deep sequence model to learn behavior sequence embedding in an unsupervised fashion, and combine them with intuitive features from risk experts to form a hybrid behavior representation; secondly,we use HDBSCAN to perform clustering on behavior features, to improve the computational efficiency, we propose a GPU accelerated version of HDBSCAN \cite{hdbscan} called \textbf{pHDBSCAN} ; thirdly, risky clusters are extracted and clustering explanation techniques are used to describe the clusters in conditional statements. We will give a detailed explanation of the algorithm in Section~\ref{sec:method}. 

To summarize, our key contributions are:
\begin{itemize}
    \item  An automatic clustering based fraud detection framework utilizing behavioral sequence data, called \textbf{FinDeepBehaviorCluster}. Based on experimental results, our proposed framework can catch fraudulent transactions missed by existing predictive risk models and significantly reduce the transaction loss.  
    \item Engineering Excellence: To address the challenge of clustering on industry-scaled data sets, we have a new implementation of GPU powered HDBSCAN (pHDBSCAN) which is several orders of magnitude faster on tens of millions of transactions. 
\end{itemize}

\section{Related Work}\label{sec:lit}

In this section, several key research areas related to our work are reviewed. 

\subsection{Click Stream Data for Fraud Detection}
Zhongfang et.al. \cite{attr_seq} proposed a framework to learn attributed sequence embedding in an unsupervised fashion, where they used encoder-decoder setting to define the attribute network, used sequence prediction setting to define the sequence network, then learned the embedding by training the integrated network, which set up a core foundation for user behavior analysis. Longfei et.al. \cite{time_attention} proposed a unified framework that combined learned embedding from users’ behaviors and static profiles altogether to predict online fraudulent transactions in a supervised fashion. Recurrent layers were used to learn the embedding of dynamic click stream data. Their proposed model managed to boost the benchmark GBDT model from $0.981$ to $0.99$ using AUC as the evaluation metric. However, they did not give a solid experimental comparison between the add-on values of different data sources from the application's view. Gang et.al..\cite{clickstream} proposed a clustering based method to analyze user behavior data for anomaly detection. Initiative features were generated to represent click streams explicitly, a similarity graph was built based on the similarity between users’ click streams. Then a divisive hierarchical clustering technique was used to iteratively partition the similarity graph. A hierarchical behavioral cluster was derived for interpretation and visualization of fraud clusters. However, handcrafted features were insufficient to represent click-stream data.  

\subsection{Clustering Techniques}

\begin{figure}
    \centering
    \includegraphics[width=\linewidth]{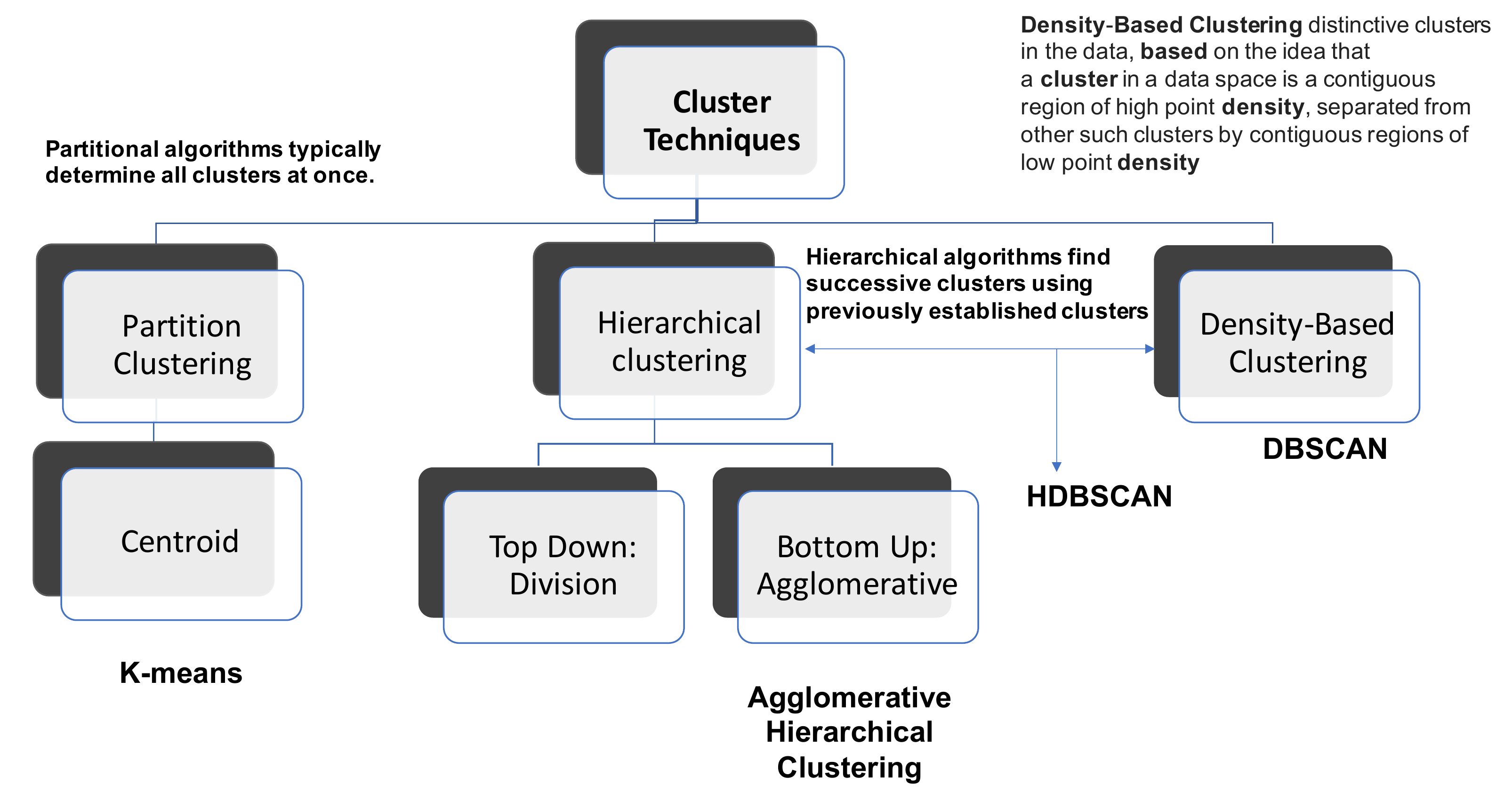}
    \caption{Commonly used Cluster Techniques}
    \label{fig:cluster}
\end{figure}

There are commonly used clustering techniques such as partitioning clustering (K-means), hierarchical clustering (Top Down: Division, and Bottom up: Agglomerative) and Density-Based clustering (DBSCAN\cite{dbscan}), illustrated in Figure~\ref{fig:cluster}. The limitations of K-means clustering include: need a pre-defined k, badly impacted by outliers, sensitive to initial seeds and not suitable for special data structures. Hierarchical clustering has lower efficiency, as it has a time complexity of $O(n^{3})$, and meanwhile is sensitive to outliers and noise. DBSCAN doesn’t perform well when the clusters are of varying densities because the epsilon parameter serves as a global density threshold which can cause challenges on high-dimensional data. Recently, Hierarchical Density-based spatial clustering of applications with noise(HDBSCAN) has become more widely used in various industries. It better fits real life data which contains irregularities, such as arbitrary shapes and noise points. The algorithm provides cluster trees with less parameters tuning, and is resistant to noise data. However, the computational cost has limited its application in large scale data sets. Especially in transaction risk domain, high performance clustering can reduce the decision time and mitigate the loss.

As mentioned earlier, clustering algorithms lead to cluster assignments which are hard to explain, partially because the results are associated with all the features in a more complicated way. While, explainable AI is a must-have in financial services, which can increase transparency and trust-worthiness of the black-box models. As a best practice in clustering tasks, decision rule generation method is used to describe each segment. Commonly used algorithms such as RuleFit\cite{rulefit} and Skope-rules\cite{skope-rules}, which aim at learning logical and interpretable rules by extracting rules from ensemble trees. While Skope-rules differ with RuleFit by using semantic deduplication rather than L1-based feature selection.



\section{Research Questions and Methodologies}\label{sec:method}
In this section, we formulate our research questions of fraudulent detection in real-world transactions. Then we propose the \textbf{FinDeepBehaviorCluster} framework for fraudulent transaction cluster mining by using user behavioral data as well as explanation generation for risky clusters. 

\subsection{Research Questions}
In this work we want to achieve the following goals: (1) an effective framework that catches fraudulent transactions missed by existing risk system; (2) a good representation of behavior sequential data for risk management; (3) an efficient clustering algorithm implementation capable of handling tens of millions transaction data within 1 hour, so that the framework is applicable in real-time risk mitigation; (4) explainable techniques which can be applied on clusters to assist Business Partners for knowledge discovery and fraud pattern mining. To be concrete, we address the following three challenges: 
\begin{enumerate}
    \item \textbf{RQ1: }how to learn a good representation of behavior sequential data?
    \item \textbf{RQ2: }how to implement a clustering algorithm with high-computational efficiency on industry-scaled data to extract risky clusters?
    \item \textbf{RQ3: }how to explain the risky clusters in a human understandable way?
\end{enumerate}

\subsection{Model Architecture }
\begin{figure}
    \centering
    \includegraphics[width=\linewidth]{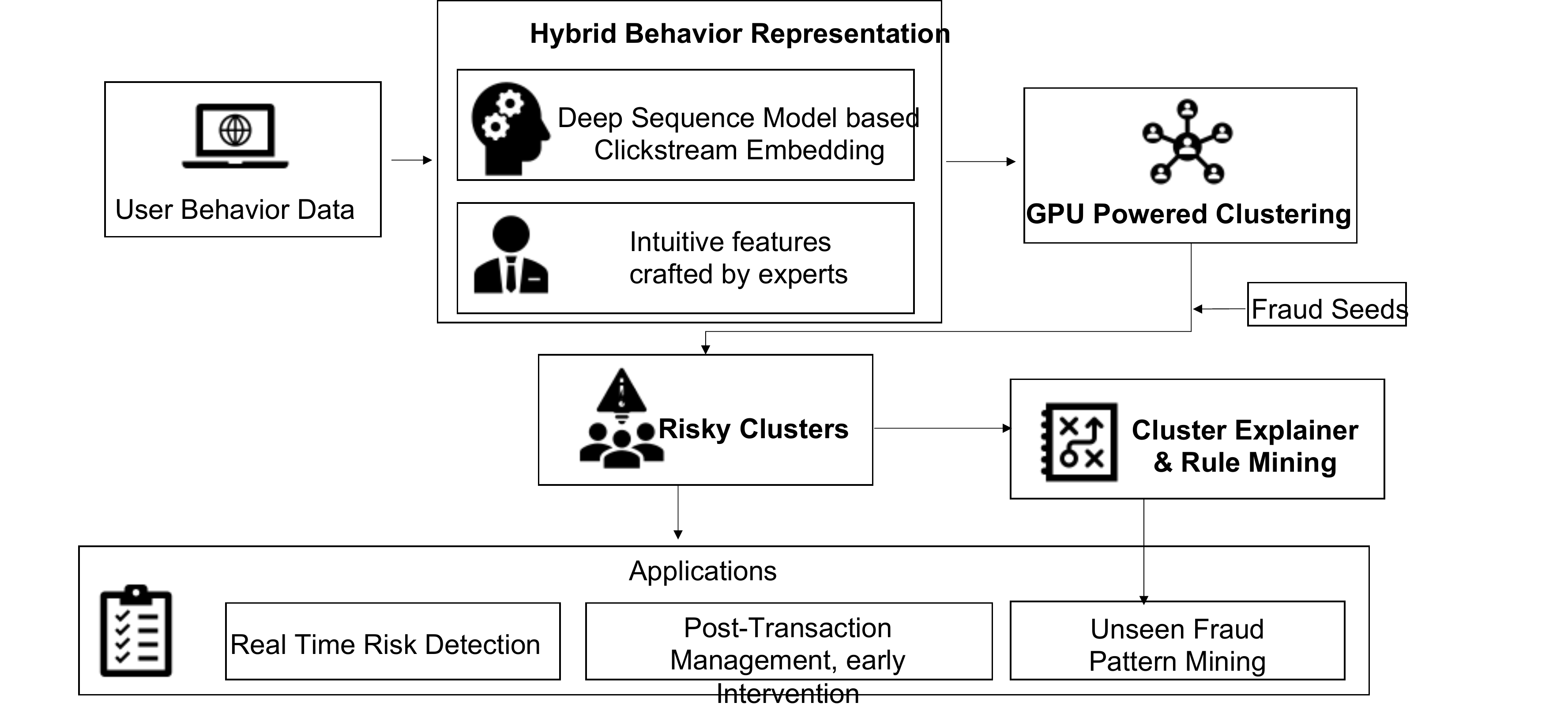}
    \caption{The proposed explainable deep learning based cluster using behavioral sequence data in transaction risk management.}
    \label{fig:framework}
\end{figure}

This work focuses on real-world automatic transaction fraud detection in e-commerce. The proposed framework works as a complement to the existing real-time risk detection systems by utilizing user behavior sequential data. It consists of 5 components illustrated in Figure~\ref{fig:framework}:
\begin{enumerate}
    \item \textbf{User behavioral data processing module:} User click-stream data collection, pre-processing and down-streaming tasks.
    \item \textbf{Hybrid Behavior Representation module:} On one hand, click stream is represented as embedding learned by advanced deep neural networks; On the other hand, intuitive risk features are created by experts for further interpretation work. Then two different paradigms of features are combined. 
    \item \textbf{GPU Powered clustering:} A proposed clustering method (\textbf{pHDBSCAN}) is implemented to achieve excellent engineering performance, and make clustering on industry scaled data possible.
    \item \textbf{Risk Cluster Generation:} Risky clusters are extracted based on heuristic rules (leveraging both fraud density and cluster's self-features such as cluster size, cluster similarity measurement, ect.) for down-stream applications.
    \item \textbf{Cluster explainer:} Cluster explainer is trained to describe the patterns of risky clusters so that the result of clustering is human-understandable. 
\end{enumerate}

The proposed framework of explainable behavioral sequence clustering has provided quite versatile risk applications. 1. Real-time risk detection: continuously identify high-risk transaction clusters using behavior data, risky clusters can be considered as negative "behavior fingerprints" essentially, new purchase orders should be blocked if they have similar user behaviors to the risky clusters. 2. Post facto, early intervention: reverse transactions if they fall in risky cluster based on preset rules. 3. Fraud pattern discovery and rule extraction: Through cluster explainer, investigate unseen fraud patterns and extract new rules to defense fraud attacks quickly. 

\subsection{Hybrid Behavior Representation}


Based on thorough data analysis, we found that fraudsters usually have some patterns linked with site behavior. For example, some fraudsters have very clear purchase intentions, they come to the site, purchase high-value items and go, while legitimate buyers usually browse a lot of items similar to the final purchase, compare prices and then place orders. Therefore, behavioral sequence is an important but usually neglected data source for fraud detection. However, mining useful information from behavioral sequence is quite challenging. In this paper, we propose a hybrid behavior sequence representation method, on one hand, intuitive features such as simple statistics of event sequence, session profile features, traffic source etc. are extracted based on domain knowledge; on the other hand, a deep learning based unsupervised model is used to learn the sequence representation. 

To identify risky behaviors, click stream data can be simplified as page events sequence, with page view and dwell time representing each single event. Inspired by the attention mechanism, we use time attention technique to handle the interaction between dwell time and page view. Say, we have a page view sequence $x$ of length $n$: $x=[x_1,x_2,…,x_n]$, $x_j$ stands for each page view, and dwell time sequence sequence $y$ of length $n$: $y=[y_1,y_2,…,y_n]$, $y_j$ stands for the dwell time user has spent on the page. Firstly, bidirectional LSTM blocks are used as an encoding part, the raw sequence $x$ will be processed to a forward hidden state $\overrightarrow{h}_{j}$ and a backward one $\overleftarrow{h}_{j}$. Similar process is applied on the dwell time sequence, with $s_{t-1}$ as the encoding hidden state. Secondly, we calculate the relationship between each page view and dwell time using $e_{t_j}=a(s_{t-1},h_j)$, where $a$ is a relationship function, here we use dot product as $a$. Thirdly, softmax is used to get the normalized attention distribution:
\begin{equation}
\alpha_{t,j} = \frac{exp(e_{tj})}{\sum_{k=1}^{n}exp(e_{tk})}
\end{equation}
Using $\alpha_{t}$ we can get weighted sum of context vector,
\begin{equation}
c_{t}=\sum_{j=1}^{n}\alpha_{t,j}h_{j}
\end{equation}
Thus, the next hidden state $s_t = f(s_{t-1},y_{t-1},c_t)$ is derived.

\begin{figure}
    \centering
    \includegraphics[width=\linewidth]{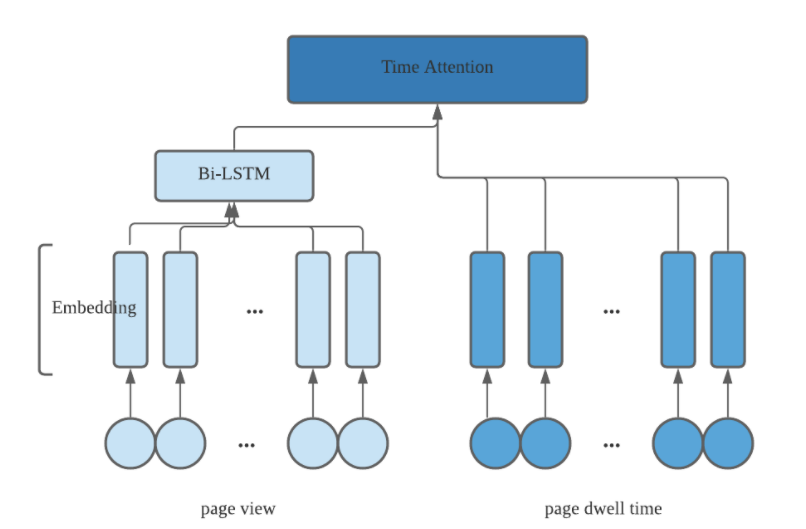}
    \caption{The proposed attention based click stream sequence embedding architecture. }
    \label{fig:attention}
\end{figure}

In our paper, the click stream sequence is modeled as Figure~\ref{fig:attention}: BiLSTM is used to process both page event sequence and page dwell time sequence, and an attention layer is put on top of the outputs of BiLSTM layer for each sequence. Fraud detection has very sparse labels, and our purpose is to discover the fraud groups with similar behavior patterns, therefore unsupervised learning mechanism is used rather than supervised fraud prediction technique, the model is trained to minimize the log likelihood of the incorrect prediction of next event. Thus the training objective function is formulated using cross-entropy as
\begin{equation}
L_{s} = -\sum_{t=1}^{l_k}\alpha_{k}^{(t)}\log{y_{k}^{(t)}}
\end{equation}

After the sequence network is trained, we use the parameters of the sequence network to embed each behavioral sequence by outputting the bottleneck layer as behavior sequence representation. 

\subsection{Clustering - GPU Powered HDBSCAN Implementation}

Our motivation to implement a GPU powered HDBSCAN is two-folded: the first part is that HDBSCAN is a clustering algorithm robust to noise in complex real world application; the second part is that the current HDBSCAN implementation \cite{hdbscan} cannot meet computational efficiency requirement on industry-scaled data sets. Inspired by K-means implementation on FAISS\cite{faiss}, we proposed a GPU power HDBSCAN (\textbf{pHDBSCAN}) comparable to existing HDBSCAN, which has optimized the distance matrix calculation, as well as the following 5 steps of HDBSCAN algorithm:

\begin{enumerate}
    \item Transform the space according to the density/sparsity.
    \item Build the minimum spanning tree (MST) of the distance weighted graph.
    \item Construct a cluster hierarchy of connected components.
    \item Condense the cluster hierarchy based on minimum cluster size.
    \item Extract stable clusters from the condensed tree.
\end{enumerate}

\subsubsection{Optimization 1: Distance matrix Calculation}

\begin{itemize}
    \item In HDBSCAN algorithm, fully pair-wised distance matrix is only used in building the minimal spanning tree (MST). However, in most situations, only k-smallest edges for each node are used to build the MST, other edges are skipped. It means that most calculation resources are not necessary in full distance matrix, instead it can be replaced by the k-nearest edge matrix.  
    \item Facebook AI Similarity Search\cite{faiss} is a super efficient library to query k-nearest neighbors. Especially, FAISS GPU accelerated version can significantly increase computational efficiency. 
    \item When $k$ is relatively small, it’s possible to speed up the query process by leveraging the K-means algorithm: a). segmenting the space into sub-spaces, b). for each $KNN$ query, search m-nearest sub-spaces firstly, then search k-nearest neighbors in m-nearest sub-spaces. This process can significantly reduce the search space. This can be done via FAISS \textit{IndexIVFFlat} Index.
    \item In summary, time complexity can be reduced from $O(|V|^2 * |dim|)$ to $O(|V| * log|V|)$. Using a $|V|=12M$, $|dim|=80$ data set, under two \textit{v100 GPUs}, it takes $15min$, where $V$ is sample size, and $dim$ is the feature dimension. 
\end{itemize}

\subsubsection{Optimization 2 -  Reachable matrix calculation}
\textit{k-core} distance can be derived from k-nearest matrix, then k-nearest matrix can be updated to reach k-nearest matrix using numpy\cite{oliphant2006guide,van2011numpy}. On a dataset with $|V|=12M$, $|dim|=80$, this process only takes about \textit{10s}.

\subsubsection{Optimization 3 -  Optimized MST algorithm}
Generally, Prim\cite{Prim} and Kruskal\cite{Kruskal} are the most common algorithms to build minimal spanning tree. Both of them take $O(m log n)$ time. We have implemented both algorithms for MST finding. Additionally, cupy\cite{cupy} was used to further speed up edge sorting in Kruskal algorithm implementation. See time efficiency comparison among different sort libraries in Table ~\ref{tab:sort_compare}.

\begin{table}[!t]
\centering
\caption{Time Efficiency comparison among different sort libraries}
\label{tab:sort_compare}
\begin{threeparttable}
\resizebox{\linewidth}{!}{
\begin{tabular}{|c|c|c|}
\hline
\textbf{Sort lib}   & \textbf{Double array size} & \textbf{Time cost} \\ \hline
Python default sort & 360M                       & $\sim$200s         \\ \hline
numpy.sort          & 360M                       & $\sim$9s           \\ \hline
cupy.sort           & 360M                       & $\sim$0.01s        \\ \hline
\end{tabular}
}
\begin{tablenotes}
  \small
  \item a) Tested on a virtual machine with 16 CPU kernels and 2 GPUs. b) Time costs for data preparation (e.g. copying data from memory to gpu) are not included.
\end{tablenotes}
\end{threeparttable}
\end{table}

\subsubsection{Optimization 4 - Build the cluster hierarchy}
When building the cluster hierarchy, Disjoint-Union set\cite{union_algo,disjoint_algo} was used as data structure to optimize the process. Disjoint-union set provides operations of adding new sets, merging sets and finding a representative member of a set. Path compression and union by rank were used when implementing the disjoint-set data structure. The total time complex is be $O(|V| * alpha(V)). alpha(v)$ , which is the inverse Ackermann function\cite{AckermannFunction}. The inverse Ackermann function grows extraordinarily slow, so this factor is $4$ or less for any $n$ in real-world situations. 

\subsubsection{Optimization 5 - Condense the cluster tree}
As real-world data set is extremely unbalanced, for instance, given $12M$ transactions, the number of samples in the deepest node can range from $500K$ to $1M$. Instead of recursive traverse, we implemented depth-first tree traversal by stacking processes to optimize the condense cluster tree procedure. 

Time efficiency comparisons for each step are provided in Table ~\ref{tab:time_effiency}


\begin{table*}[!ht]
\caption{Time efficiency comparisons between pHDBSCAN and HDBSCAN}
\label{tab:time_effiency}
\centering
\begin{tabular}{|c|c|c|c|}
\hline
\textbf{Step}                      & \textbf{Vanilla HDBCAN}     & \textbf{pHDBSCAN} & \textbf{Proposed Optimization Solution} \\ 
\hline
1, Calculate the distance matrix        & $O(|V|^{2} * |dim|)$ & $O(|V| * log|V|)$           & Optimization 1          \\ 
\hline
2, Transform the space             & $O(|V|^2)$         & $O(|V|)$                    & Optimization 2          \\ 
\hline
3, Build the minimal spanning tree & $O(|E| * log|V|)$                   & $O(|V| * log|V|)$          & Optimization 3          \\ 
\hline
4, Build the cluster hierarchy     & $O(|V| * log|V|)$                  & $O(|V| * Arcer|V|)$         & Optimization 4          \\ 
\hline
5, Condense the cluster tree       & $O(N)$                              & $O(N)$                      & Optimization 5          \\ 
\hline
6, Extract the clusters            & $O(N)$                              & $O(N)$                      &                          \\ 
\hline
\end{tabular}
\end{table*}

\subsection{Clustering Explainability }

\begin{figure}
    \centering
    \includegraphics[width=\linewidth]{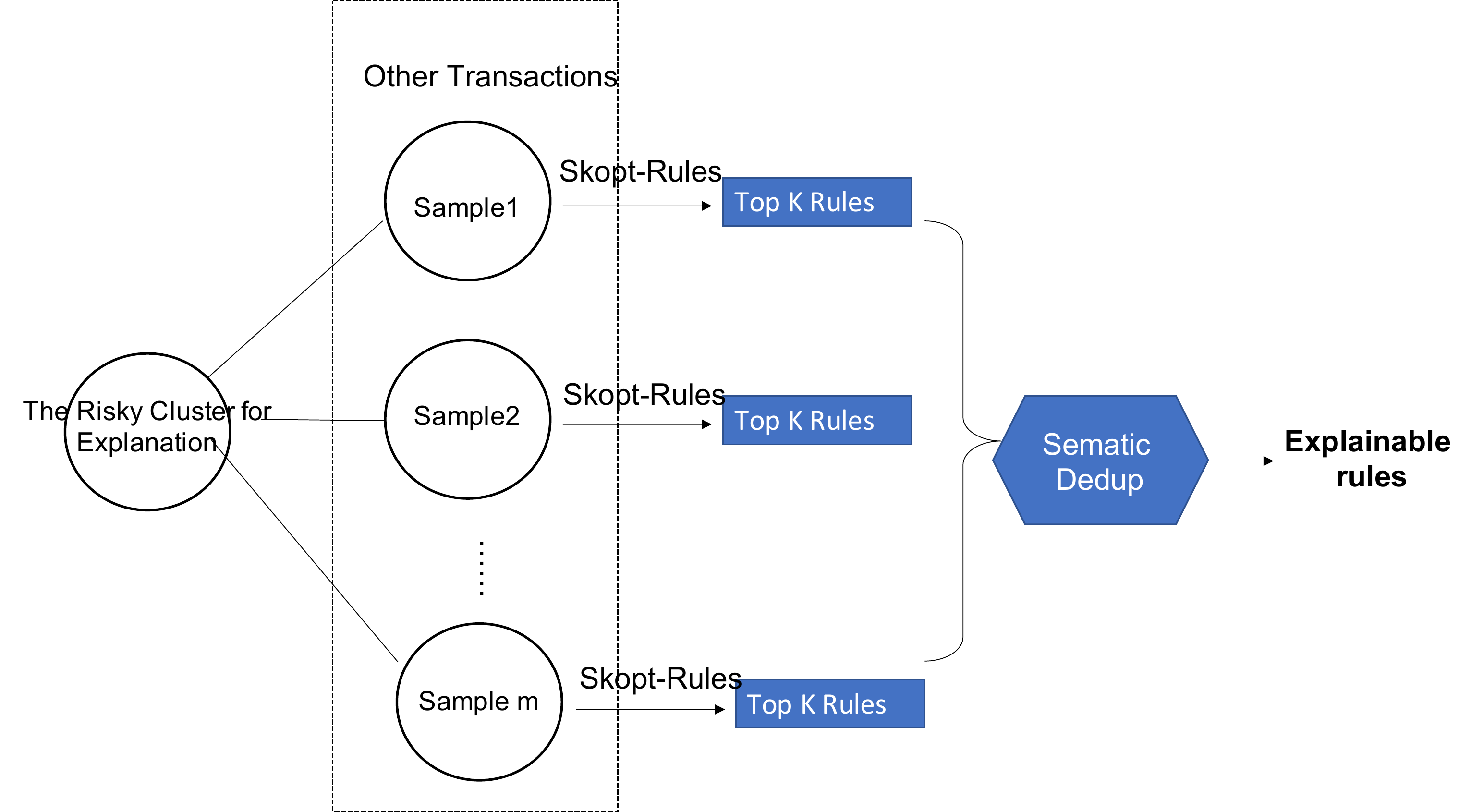}
    \caption{Click Stream Visualization for Risky Clusters}
    \label{fig:rule_explain}
\end{figure}

In the proposed \textbf{FinDeepBehaviorCluster} framework, we use rule extraction algorithm to describe the characteristics of detected risky clusters for explanation. Intuitive features $x$ are used to predict the target $y[0,1]$, with $1$ indicating the transaction falls into risky clusters for explanation, and $0$ means the transaction is not from risky clusters. We use Skope-Rule as our rule extraction tool, which is a trade off between interpretability of decision trees and predicting power of bagging trees. It firstly generates a set of logical rules using bagging trees, then keeps only high performance rules given predefined precision/recall threshold, finally applies semantic rule deduplication and drops homogeneous rules. To increase the robustness of explanation results, we use a bootstrap mechanism and sample different batches for negative examples. See the set-up in ~Figure:\ref{fig:rule_explain}.

\section{Experimentation:  Clusters Algorithm }
\subsection{Compare pHBDSCAN with HDBSCAN on Synthetic Data sets}
\begin{figure}
    \centering
    \includegraphics[width=\linewidth]{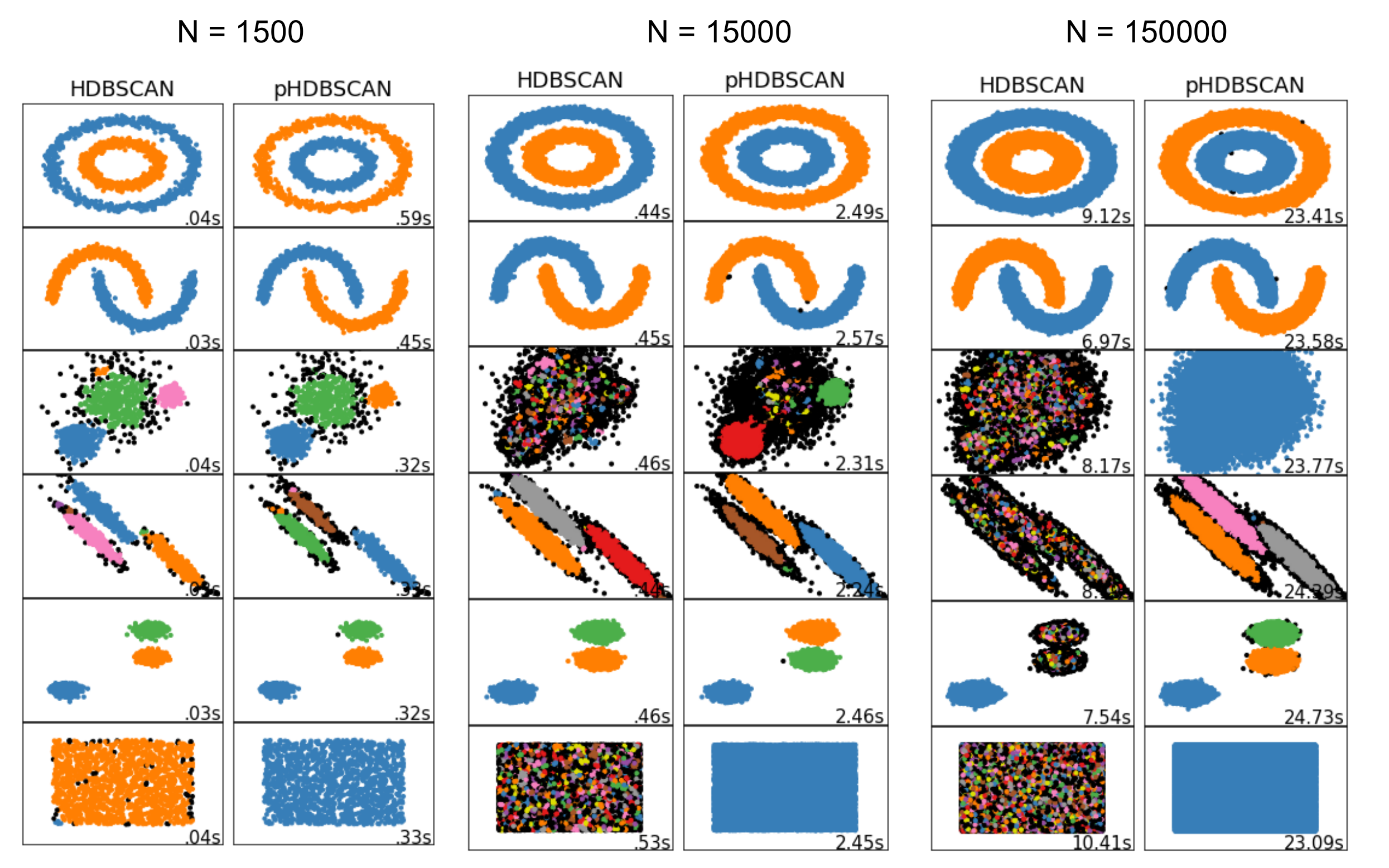}
    \caption{Intuitive comparisons between HDBSCAN and pHDBSCAN}
    \label{fig:cluster_sim}
\end{figure}

To give an intuitive impression of how pHDBSCAN performs on data sets with varied interesting structures, experiments were carried out using various sklearn cluster algorithms for comparison\cite{plotClusterComparison}. In our experiments, data generation parameters are the same as those of sklearn, in addition, sample sizes are varied to show how algorithms perform on different volumes of datasets. See 2D plots in Figure~\ref{fig:cluster_sim}.

\subsection{Compare HDBSCAN \& pHBDSCAN on Benchmark Data sets}
\begin{figure}
    \centering
    \includegraphics[width=\linewidth]{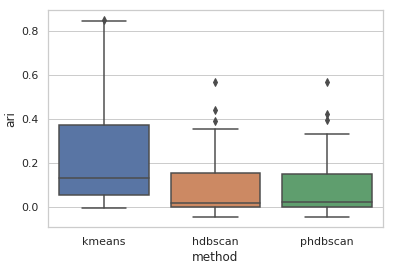}
    \caption{Comparisons with HDBSCAN and pHDBSCAN on real world cluster benchmark data sets}
    \label{fig:cluster_compare}
\end{figure}

In this section, the performance of HBDSCAN and pHDBSCAN are compared on public benchmark data sets. Given the ground truth of class labels for clustering tasks, it is common to use Adjusted Rand Index (ARI) \cite{ARI} to evaluate the performance of clustering algorithms, which measures the similarity between two assignments. We use clustering benchmark data sets with ground truth of classes contributed by Tomas et al.\cite{cluster_benchmark_data}. Because we are solving an industry problem, the $22$ valid real-world benchmark data sets are used. Some statistics of the data sets: sample size: $min = 101$, $max = 20000$, $median = 343$; number of features: $min = 2$, $max = 262$, $median = 10$; number of classes: $min = 2$, $max = 116$, $median = 3$. From the result in Figure ~\ref{fig:cluster_compare}, although both HDBSCAN and pHDBSCAN have lower performance compared with K-means given the ground truth of $K$, pHDBSCAN has a comparable result to vanilla HDBSCAN.

\section{Experimentation: Fraud Detection in Transaction Risk }

\subsection{Data and prepossessing}
\begin{figure}
    \centering
    \includegraphics[width=\linewidth]{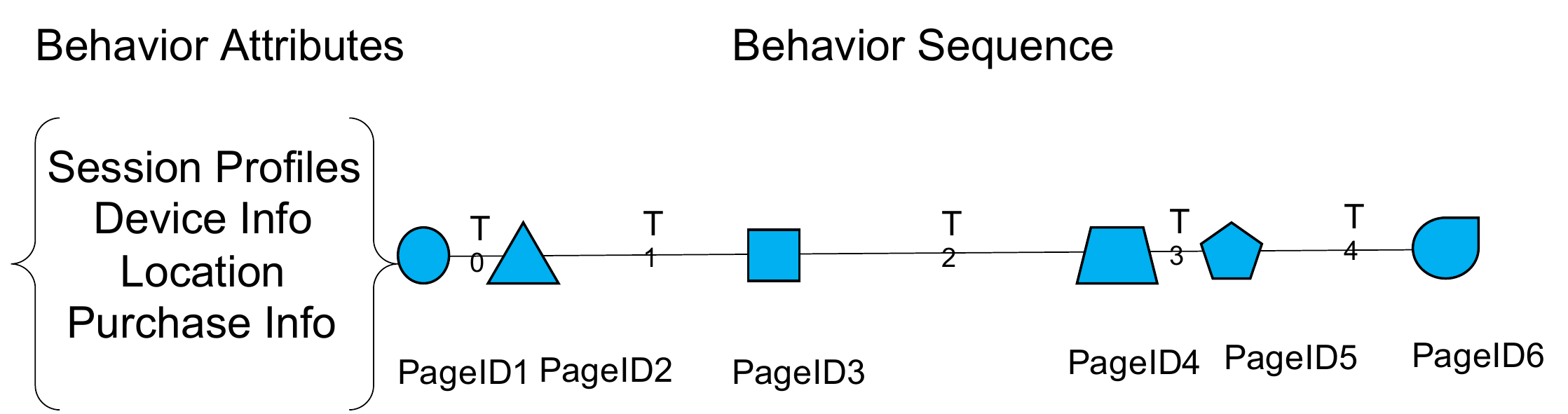}
    \caption{Illustration of click stream data with attribute.}
    \label{fig:beh_data}
\end{figure}

In our experiments, transaction behavioral data is comprised of two parts: behavior attributes and behavior sequence. Behavior attributes include session level profiles (channel, traffic referrer, session level characteristics, etc.), device information (OS type, screen resolution, etc.), location (IP etc.) and purchase information (item price, item category, ect. ). Behavior sequence is click-stream data on e-commerce site, to be specific, page view types and dwell time on pages. See the data formats in Figure~\ref{fig:beh_data}. 

Risky transactions are used as seeds for risky cluster generation. Risk transactions refer to confirmed fraudulent transactions, transactions declined by payment processors. Confirmed fraud transactions are delayed feedback, while declined transactions are real-time responses. Leveraging both types of feedback can improve accuracy and time efficiency. 

\subsection{Experimentation Design for Transaction fraud Detection}
\textbf{Experimentation Design}: We designed two experiments for different purposes. Firstly, as a key component of the framework, \textbf{pHDBSCAN} is compared with other mainstream clustering algorithms in both engineering efficiency and model accuracy of a predictive framework. Secondly, the framework is applied on transaction fraud detection tasks from the view of post-transaction management. Both experiments are run on real-world e-commerce transaction data sets. 

\noindent \textbf{Evaluation Metric}: Unlike supervised learning, unsupervised learning like clustering does not have standard algorithm-level evaluation metrics, instead metrics from downstream business applications are used for clustering evaluation. In fraud detection tasks, we usually use precision, recall, return rate (Loss Saved $/$ Profit Hurt) as metrics.

\subsubsection{\textbf{Inductive Experimentation}} 

Compare time efficiency and model performance among different cluster techniques. To simplify the problem, the comparison among different clustering techniques is designed as an induction framework , see Figure~\ref{fig:exp_inductive}. Transaction data is split into training data set and testing data set chronologically. Clustering algorithms are run on training data, risky clusters are extracted based on heuristics rules, such as cluster size, coherence confidence, fraud density, etc.; When a new transaction happens, cluster id is assigned to this transaction indicating which cluster it belongs to; If the transaction is assigned to risky clusters, it is predicted as fraudulent. For the algorithm without prediction function, KNeighborsClassifier is used to assign cluster labels, with parameters $\text{\textit{n\_neighbors}} = 5$, $\text{\textit{weights}} = 'distance'$. The result can be found in Table \ref{tab:inductive_expt}. Compared with vanilla HDBSCAN, pHDBSCAN increases computational efficiency by more than 500 times, and meanwhile has better model performance. More detailed experimentation set-up is as follows:

\begin{figure}
    \centering
    \includegraphics[width=\linewidth]{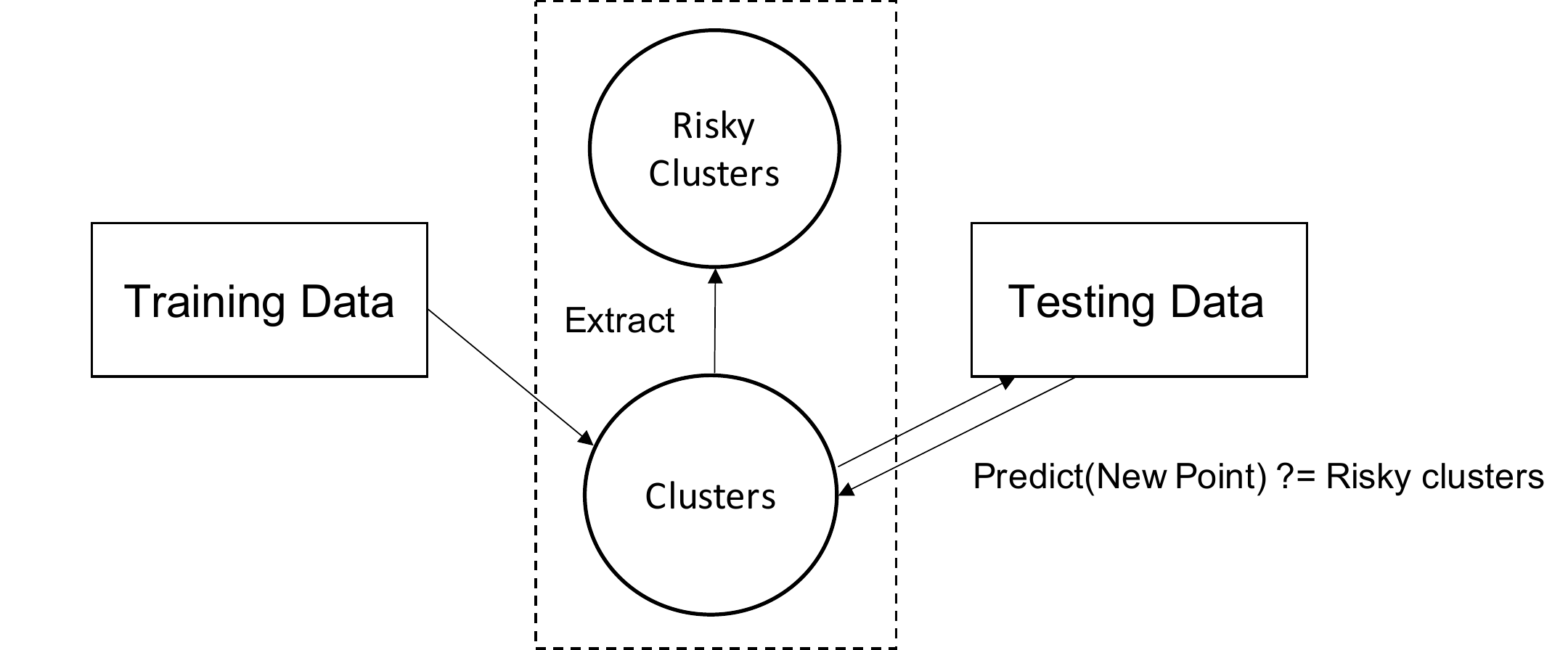}
    \caption{The inductive experimentation set up in fraud detection.}
    \label{fig:exp_inductive}
\end{figure}

\begin{itemize}
    \item Dataset: only behavioral sequence embedding with 80-D is used as clustering features, with 0.4M samples in training set and 0.1M in testing set.
    \item Platform: $\text{\textit{cpu}} : 32$, $\text{\textit{memory}} : 128$, $\text{\textit{GPU}} : 1-tesla-m40$
\end{itemize}
A few key parameters of different clustering Strategies:
\begin{itemize}
    \item HDBSCAN: Implemented by python library HDBSCAN\cite{hdbscan}.
    \item Sequence Dimension Reduction HDBSCAN: for PCA, $\text{\textit{d\_in}} = 80$, $\text{\textit{out\_d}} = 20$, $\text{\textit{eigen\_power}} = -0.5$, implemented by FAISS\cite{faiss}, and vanilla python version HDBSCAN.
    \item OPTICS: Algorithm implemented by sklearn \cite{OPTICS_sklearn,OPTICS_paper}.
    \item GPU KMeans: Implemented by FAISS, with $\textit{best\_ncentroids} = 12000$, $\textit{max\_niter } = 100$, nearest centroid used for inference.
    \item pHDBCAN: Our proposed GPU power HDBSCAN implementation.
    
\end{itemize}

\begin{table}[!t]
\centering
\caption{Performance comparisons among different cluster techniques in FinDeepBehaviorCluster framework}
\label{tab:inductive_expt}
\begin{threeparttable}
\resizebox{\linewidth}{!}{
\begin{tabular}{|c|c|c|c|c|c|}
\hline
 Strategy                                 & time\_cost & \ \#.Risky C & \textbf{Precision} & \textbf{Recall} & \textbf{F-score} \\ \hline
HDBSCAN                       & $\sim$2days      & 31                 & 8\%                & 0.20\%          & 0.38\%           \\ \hline
PCA(20)-\textgreater{}HDBSCAN & 3hours           & 25                 & 5\%                & 0.12\%          & 0.23\%           \\ \hline
OPTICS                        & 1$\sim$2 days    & 45                 & 4\%                & 0.17\%          & 0.32\%           \\ \hline
Kmeans(K=12000)               & 5mins* Niters    & 25                 & 14\%               & 0.58\%          & 1.12\%           \\ \hline
\textbf{pHDBSCAN}             & \textbf{5mins}   & \textbf{37}        & \textbf{17\%}      & \textbf{0.39\%} & \textbf{0.75\%}  \\ \hline
\end{tabular}
}
\begin{tablenotes}
  \small
  \item Metrics reported on sampled traffic.
\end{tablenotes}
\end{threeparttable}
\end{table}




\subsubsection{\textbf{Transductive Experimentation}}

\begin{figure}
    \centering
    \includegraphics[width=\linewidth]{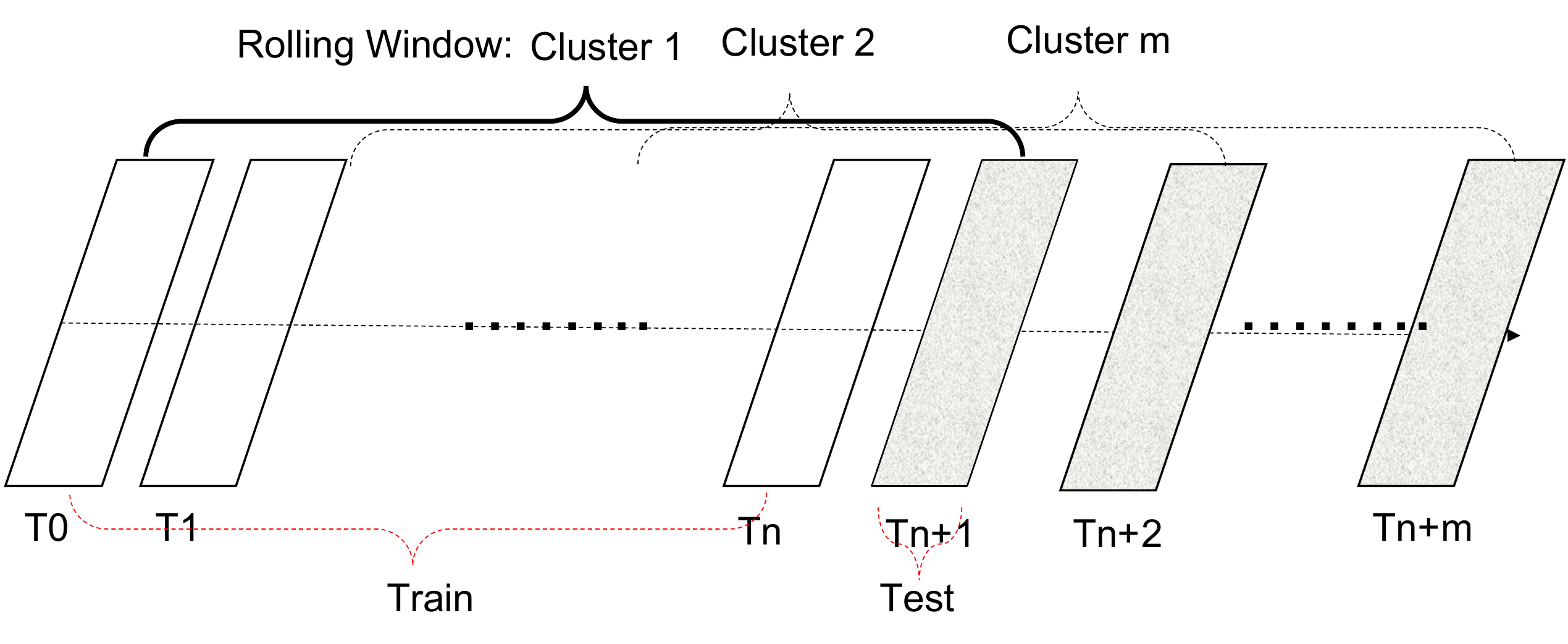}
    \caption{The proposed transductive experimentation setup for fraud transaction.}
    \label{fig:exp_transductive}
\end{figure}

This experiment is set up as a transductive learning framework run on rolling time windows, see Figure~\ref{fig:exp_transductive}. For each clustering batch, transactions from the first $n$ snapshots are used as training data, with the $n+1$ snapshot transactions as testing data. Each transaction from the testing period is determined as fraudulent or not by examining whether it belongs to a risky cluster. To increase the proportion of fraudulent transactions, stratified sampling technique is applied on training data considering both label distribution and time decay effectiveness. In this experiment, we compare each component of behavior feature representation, and find that sequence embedding is a good representation of click stream data, furthermore, combining sequence embedding with handcrafted features can achieve the best performance for fraudulent transaction detection. From the results showed in Table: \ref{tab:transductive_expt}, we can see the proposed framework \textbf{FinDeepBehaviorCluster} can effectively catch missed fraudulent transactions with the highest profitable return rate.

\begin{table*}[!ht]
\centering
\caption{Compare performance on behavior feature representations using pHDBSCAN}
\label{tab:transductive_expt}
\resizebox{\linewidth}{!}{
\begin{tabular}{|c|c|c|c|c|c|c|c|c|}
\hline
Behavior. Ftr  & Dim       & \#.Risky Cluster & Precision        & Recall          & F-score         & LossSave                     & ProfitHurt      & ReturnRate     \\ \hline
Hand Ftr            & 5M*67D           & 30                 & 18.27\%          & 0.92\%          & 1.75\%          &\$3335.53 & \$1492.32         & 2.24           \\ \hline
Deep Seq Ftr   & 5M*80D           & 19                 & 70.46\%          & 1.71\%          & 3.35\%          & \$6214.81                      & \$260.56          & 23.85          \\ \hline
\textbf{Hybrid Ftr} & \textbf{5M*147D} & \textbf{25}        & \textbf{70.17\%} & \textbf{3.28\%} & \textbf{6.27\%} & \textbf{\$11895.58}            & \textbf{\$505.72} & \textbf{23.52} \\ \hline
\end{tabular}
}
\begin{tablenotes}
  \small
  \item Metrics reported on sampled traffic.
\end{tablenotes}
\end{table*}

\section{Case Studies: Risky Cluster Understanding}

\begin{figure}
    \centering
    \includegraphics[width=\linewidth]{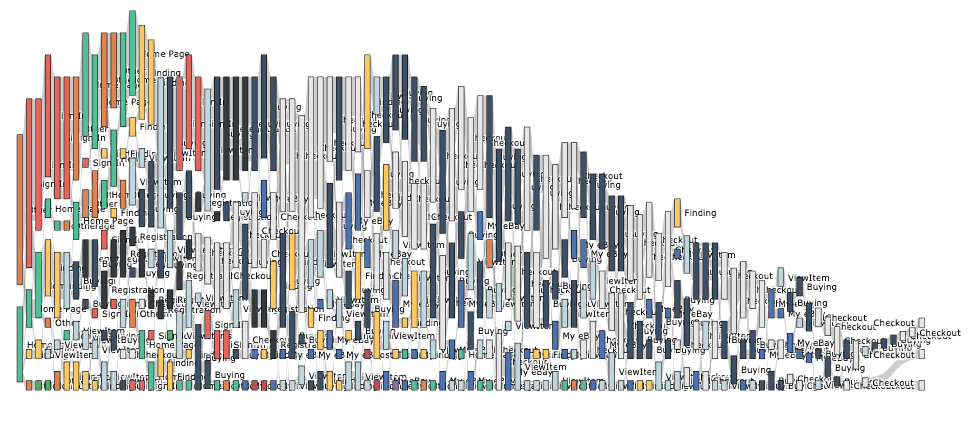}
    \caption{Click Stream Visualization of Risky Cluster }
    \label{fig:case1995}
\end{figure}

Following the proposed \textbf{FinDeepBehaviorCluster} framework, risky clusters are extracted and explained in a rule-based, human understandable manner. Let's review a risky cluster for pattern mining and domain knowledge study. See the click stream visualization using Sankey diagram in Figure ~\ref{fig:case1995}.

Using the cluster explanation method, top performing rules are extracted. The example of one rule is listed as follow for illustration:

\begin{quote}
\begin{small}
$number\_checkout\_events > 10$ \\
$and \quad search\_counts < 1$\\
$and \quad dwell\_time\_on\_view\_item < 5$\\
$and \quad average\_sign\_up\_time < 2$\\
$and \quad payment\_method\_creditCard = 1$\\
$and \quad order\_amt > 150$ \\
$and \quad item\_category = 'Electronic'$\\
$and \quad email\_domain = 'qq.com'$\\

\end{small}
\end{quote}

By carefully reviewing these rules, our business partner has identified that the cluster belongs to 'repeated offender' with the following characteristics:

\begin{itemize}
    \item Newly registered buyers or guests.
    \item Use a stolen financial instrument.
    \item Multiple transactions in one session, even multiple accounts registered in one single session.
    \item Very clear shopping targets: resell-able and high value product purchases, such as high-end fashion and electronic items.
    \item Very few search or item browsing events
    \item Very familiar with the site: act in a very high speed, and proceed smoothly. 
    \item Similar behavior attributes, for example, similar patterns of registration email, traffic source, device type, et al. 
\end{itemize}

\section{Conclusion}
In this paper, we propose \textbf{FinDeepBehaviorCluster}, a systematic way of utilizing click-stream data for fraud detection and fraud pattern mining. Specifically, time attention based Bi-LSTM is used to learn the embedding of behavior sequence data. In addition, to increase the interpretability of the system, handcrafted features are generated to reflect domain knowledge. Combing the two types of features, a hybrid behavior representation has formed. Then a GPU optimized HDBSCAN algorithm called \textbf{pHDBSCAN} is used for clustering transactions with similar behaviors. Risky clusters are extracted using heuristic rules. Last, fraudulent transactions are determined for various application purposes. We conduct experiments on two real-world transaction data sets and show that the proposed pHDBSCAN has achieved comparable performance to vanilla HBDSCAN, but with hundreds of times of computation efficiency increase; Our proposed \textbf{FinDeepBehaviorCluster} framework can catch missed fraudulent transactions with a decent business return rate. We also show a real-world case study where cluster explanation techniques are utilized to generate human understandable rules with high precision and recall, which facilitates further understanding and decision-making for business units in risk behavioral patterns mining. As a future work, quantitative evaluation of cluster explanation techniques will be studied. And the effectiveness of this framework with better utilizing unstructured behavior sequence data in other marketplace risk scenarios will be comprehensive validated.

\bibliographystyle{aaai}
\bibliography{ref}

\end{document}